\documentclass[conference]{IEEEtran}
\IEEEoverridecommandlockouts

\usepackage{cite}
\usepackage{amsmath,amssymb,amsfonts}
\usepackage{algorithmic}
\usepackage{graphicx}
\usepackage{textcomp}

\usepackage{comment}
\usepackage{multirow}
\usepackage{threeparttable}
\usepackage{booktabs}
\usepackage{multirow}
\usepackage[table, dvipsnames]{xcolor}
\usepackage{colortbl}
\usepackage{enumitem}
\usepackage{multirow, threeparttable, booktabs, makecell}
\usepackage{adjustbox}
\usepackage{CJKutf8}

\usepackage[hidelinks]{hyperref}  

\def\BibTeX{{\rm B\kern-.05em{\sc i\kern-.025em b}\kern-.08em
    T\kern-.1667em\lower.7ex\hbox{E}\kern-.125emX}}
\begin{document}

\title{PARCO: Phoneme-Augmented Robust Contextual ASR via Contrastive Entity Disambiguation
}

\author{\IEEEauthorblockN{Jiajun He \thanks{This work was done during Jiajun He's internship at CyberAgent.}}
\IEEEauthorblockA{\textit{Graduate School of Informatics} \\
\textit{Nagoya University}\\
Nagoya, Japan \\
0009-0006-6489-5220}
\and
\IEEEauthorblockN{Naoki Sawada}
\IEEEauthorblockA{\textit{AI Lab} \\
\textit{CyberAgent}\\
Tokyo, Japan \\
0009-0003-5883-0859}
\and
\IEEEauthorblockN{Koichi Miyazaki}
\IEEEauthorblockA{\textit{AI Lab} \\
\textit{CyberAgent}\\
Tokyo, Japan \\
0000-0002-5796-4535}
\and
\IEEEauthorblockN{Tomoki Toda}
\IEEEauthorblockA{\textit{Information Technology Center} \\
\textit{Nagoya University}\\
Nagoya, Japan \\
0000-0001-8146-1279}
}

\maketitle


\begin{abstract}

Automatic speech recognition (ASR) systems struggle with domain-specific named entities, especially homophones. Contextual ASR improves recognition but often fails to capture fine-grained phoneme variations due to limited entity diversity. Moreover, prior methods treat entities as independent tokens, leading to incomplete multi-token biasing. To address these issues, we propose Phoneme-Augmented Robust Contextual ASR via COntrastive entity disambiguation (PARCO), which integrates phoneme-aware encoding, contrastive entity disambiguation, entity-level supervision, and hierarchical entity filtering. These components enhance phonetic discrimination, ensure complete entity retrieval, and reduce false positives under uncertainty. Experiments show that PARCO achieves CER of 4.22\% on Chinese AISHELL-1 and WER of 11.14\% on English DATA2 under 1,000 distractors, significantly outperforming baselines. PARCO also demonstrates robust gains on out-of-domain datasets like THCHS-30 and LibriSpeech.

\end{abstract}


\vspace{-1mm}
\section{Introduction}
\vspace{-1mm}
\label{sec:intro}
\begin{CJK}{UTF8}{gbsn}

In recent years, end-to-end (E2E) automatic speech recognition (ASR) system has achieved high accuracy in transcribing speech \cite{kheddar2024automatic,zhang2024cif, xu2024dynamic}. However, even the most advanced ASR systems still struggle with errors in recognizing domain-specific named entities, such as technical terms and place names \cite{huber2021instant}. These ASR errors significantly affect the performance of downstream tasks such as video summarization \cite{yang2024multi, he24_interspeech} and emotion recognition \cite{he2024mf, li2024speech, shi2024study, tian2023semi, mi2024two, he25_interspeech}.


To address this challenge, researchers have developed a variety of contextual biasing methods that incorporate external knowledge to improve named entity (NE) recognition. Such contextual information is often derived from sources like lecture videos, meeting transcripts, and user contact lists \cite{sun2022tree, yang2024mala, huang2024improving}. These methods can be broadly categorized into four types:
(1) Shallow fusion approaches, which integrate auxiliary language models during decoding to adjust outputs based on context \cite{williams2018contextual, le2021deep, chen2022factorized}. Although effective, these methods typically require careful parameter tuning and rely on predefined contextual prefixes.
(2) Trie-based deep biasing, which enhances decoding constraints by organizing contextual vocabulary into a trie structure. Although efficient for lookup and constraint enforcement, this approach often requires model fine-tuning for each target domain.
(3) ASR error correction (AEC) methods, which post-process transcriptions without retraining the base ASR model \cite{he2023ed, he2024mf, li2024crossmodal, he2023enhancing, he2025pmf}. However, they usually demand domain-specific AEC models and tend to lack robustness across domains.
(4) Context-integrated training, which incorporates contextual knowledge directly into the text encoder and jointly trains it with the ASR model \cite{huber2021instant, pundak2018deep, han2022improving, liu2022internal, zhang2022end, munkhdalai2023nam+, zhou2023copyne, fang2025improving, fang2025token}, enabling E2E context-aware recognition.

Although existing methods have made progress in improving entity recognition, they still encounter notable limitations.
One key issue is that most models treat entities as sequences of independent tokens, thereby ignoring their holistic structure as unified semantic units.
As a result, multi-token entities are frequently decoded as fragmented or incomplete spans, undermining the effectiveness of biasing strategies and leading to partial or erroneous entity reconstruction.
Second, distinguishing phonetically similar entities remains difficult, and the generalization ability for rare words is limited in low-resource scenarios. To address this issue, existing studies often introduce phoneme features to enrich entity representations \cite{bruguier2019phoebe, chen2019joint, pandey2023procter, futami2024phoneme}. Phoneme information helps mitigate ambiguities caused by similar-sounding words, making it a promising direction for improving entity recognition. Typically, simple sequence encoders are used to extract phoneme representations, which are then combined with text features to reduce confusion.

However, existing phoneme encoders often fail to fully exploit fine-grained phonetic distinctions among similar-sounding entities and lack explicit disambiguation mechanisms during decoding. 
Moreover, they typically treat phoneme information as auxiliary input and do not integrate it into the biasing mechanism in a discriminative way. This limitation becomes particularly evident in open-domain conditions, where attention-based entity retrieval may still mistakenly attend to phonetically similar but semantically irrelevant candidates.


To overcome these challenges, we propose \textbf{PARCO}—a \textbf{P}honeme-\textbf{A}ugmented \textbf{R}obust contextual biasing method via \textbf{CO}ntrastive entity disambiguation. PARCO is a novel biasing-aware ASR framework that incorporates fine-grained phoneme-aware encoding and multiple decoding-time enhancements for robust and accurate entity recognition. 
Specifically, PARCO integrates a contrastive entity disambiguation (CED) loss to explicitly enhance the decoder’s ability to distinguish between phonetically similar entities, improving robustness against confusable pronunciations. To ensure accurate recognition of multi-token entities, we further introduce an entity-level supervision mechanism that encourages the decoder to attend to complete entity spans rather than making token-wise predictions. Moreover, a hierarchical entity filtering (HEF) strategy is employed to dynamically refine the biasing list based on phoneme-level similarity and confidence gating, effectively reducing false positives under ambiguous conditions.
Our main contributions are as follows:
\vspace{-1mm}
\begin{itemize}[leftmargin=*]
\setlength{\itemsep}{0pt}
\setlength{\parsep}{0pt}
\setlength{\parskip}{0pt}

\begin{figure*}[!t]
  \centering
  \includegraphics[width=2\columnwidth]{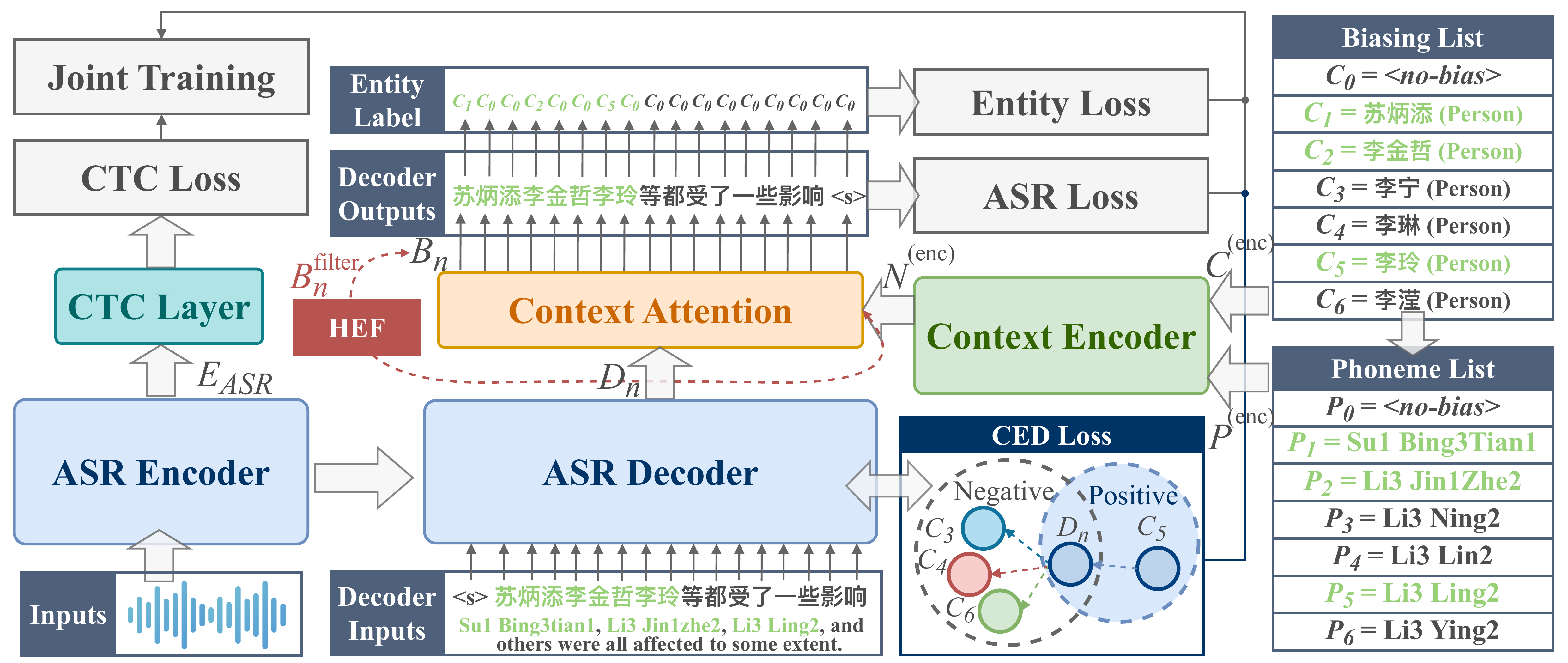}
    \caption{Overall architecture of the proposed PARCO model. The ASR backbone is based on a Conformer encoder-decoder architecture. A biasing list containing multi-token entities is encoded by a phoneme-enriched text encoder to support robust contextual biasing. The contrastive entity disambiguation (CED) loss enhances discriminability among phonetically similar entities and is detailed in Section~\ref{sec:ced}. The hierarchical entity filtering (HEF) strategy, used only during inference, dynamically refines the biasing list for improved precision under ambiguity, as described in Section~\ref{sec:hef}.
}
  \label{fig:model}
\end{figure*}

\item We propose a phoneme-augmented biasing model with a novel CED loss, which enhances the discriminative power for phonetically similar entities.
\item We design an entity-level loss mechanism to ensure the complete recognition of multi-token biased entities, avoiding partial selection during decoding.
\item We introduce a HEF strategy that combines phoneme-aware similarity retrieval and confidence-based gating during inference, improving both precision and recall of entity biasing.
\item We conduct extensive experiments on both Chinese (AISHELL-1, THCHS-30) and English (DATA2, LibriSpeech) benchmarks. The results demonstrate that PARCO significantly outperforms prior biasing methods in both in-domain and out-of-domain (OOD) scenarios.


\end{itemize}

\vspace{-2mm}
\section{AED ASR Model}
\vspace{-1mm}
An attention-based encoder decoder (AED) ASR model typically consists of an audio encoder and an attention-based decoder. The encoder transforms the input audio sequence 
\( S \) 
into an hidden representation 
\( E_{\text{ASR}} = (e_1, \dots, e_M) \in \mathbb{R}^{M \times d} \), 
where $M$ the length of the encoded feature sequence and \( d \) is the feature dimension.
At decoding step \( n \), the decoder generates an hidden representation 
\( D_{n} \in \mathbb{R}^{d} \) 
based on the previously decoded tokens 
\( T_{1:n-1} = (t_1, \dots, t_{n-1}) \) 
and \( E_{\text{ASR}} \). The probability of the predicted token \( \hat{t}_n \) is computed as:
\begin{equation}
\label{eq:1}
\hat{t}_n = \text{Softmax}( D_{n} W_n + b_n),
\end{equation}
\vspace{-1mm}where \( W_n \in \mathbb{R}^{d \times V} \) and \( b_n \in \mathbb{R}^{V} \) are learnable parameters, and \( V \) is the vocabulary size. Here, \( \hat{t}_n \) represents \( p(t_n | T_{1:n-1}, S) \), the probability distribution over possible tokens. The model is trained by minimizing the negative log-likelihood:
\vspace{-1mm}
\begin{equation}
\mathcal{L}_{\text{ASR}} = -\sum_{n=1}^{N} \log p(t_n | T_{1:n-1}, S).
\end{equation}
\vspace{-1mm}
Additionally, the hybrid CTC/attention model~\cite{kim2017joint, moriya2020self} incorporates a CTC loss $\mathcal{L}_{\text{CTC}}$ to enhance efficiency and performance.

\vspace{-1mm}
\section{Proposed PARCO Method}
As shown in Fig. \ref{fig:model}, PARCO consists of an AED ASR model, a context encoder, and a context attention module, enhancing biasing for phonemically similar entities.

\vspace{-2mm}
\subsection{Context Encoder}
\vspace{-1mm}


\noindent \textbf{Text Encoder} embeds the biasing list \( C = (C_1, C_2, \dots, C_L) \in \mathbb{R}^{L} \) with an additional ``$<$no-bias$>$" token ($C_0$), following \cite{huber2021instant}. Each entity \( C_l \) is encoded using stacked LSTM layers:  
\begin{equation}
C^{{\rm (enc)}}_{l} = \text{LSTM}(C_l) \in \mathbb{R}^{d}.
\end{equation}  
The final text representations are given by: 
\begin{equation}
C^{{\rm (enc)}} = (C^{{\rm (enc)}}_0, C^{{\rm (enc)}}_1, \dots, C^{{\rm (enc)}}_L) \in \mathbb{R}^{(L+1) \times d}.
\end{equation}

\noindent \textbf{Phoneme Encoder} embeds the phoneme sequence of an entity, denoted as $P = (P_1, P_2,\cdots , P_L) \in \mathbb{R}^{L}$, into a sequence of phoneme representations \( P^{{\rm (enc)}} = (P^{{\rm (enc)}}_0, P^{{\rm (enc)}}_1, \ldots, P^{{\rm (enc)}}_L) \in \mathbb{R}^{(L+1) \times d} \). We adopt stacked LSTM layers to capture the contextualized phoneme representations for each entity.




Finally, we concatenate \( C^{{\rm (enc)}} \) and \( P^{{\rm (enc)}} \), and pass them through a linear layer to obtain the entity representations enriched with phoneme information \( N^{{\rm (enc)}} = (N^{{\rm (enc)}}_0, N^{{\rm (enc)}}_1, \ldots, N^{{\rm (enc)}}_L) \in \mathbb{R}^{(L+1) \times d} \):


\begin{equation}
N^{{\rm (enc)}} = (C^{{\rm (enc)}} \oplus P^{{\rm (enc)}}) W_{N} + b_{N},
\end{equation}
where \( W_n \in \mathbb{R}^{2d \times d} \) and \( b_n \in \mathbb{R}^{d} \) are the learned parameters, and ``$\oplus$" denotes a concatenate function.

\subsection{Context Attention}
We calculate the attention score for entity \( C_l \) at step \( n \), using \( D_{n} \) as the query and the phoneme enriched text representation of the entity as the key \( N^{{\rm (enc)}} \):
\vspace{-2mm}
\begin{equation}
\label{eq:6}
  s_{n,l}   = \text{Softmax}\left(\frac{{D_{n} W_Q \cdot (N_l^{{\rm (enc)}} W_K)^\top}}{{\sqrt{d}}}\right) \in \mathbb{R}^{L+1},
\end{equation}
where $W_Q$, $W_K\in \mathbb{R}^{d \times d_h} $ are learned parameters and $d_h$ is the attention dimension. $ s_{n,l}$ also represents $p_c(C_l | T_{1:n-1}, S)$, which is the selection probability for the entity.
Then, we use $N^{{\rm (enc)}}$ as the value, as the bias vector passed to the ASR decoder should primarily contain textual information, to obtain the final biasing representation $B_{n}$:
\begin{equation}
\label{eq:7}
B_{n} = \sum_{l=0}^{L}   s_{n,l} N_l^{{\rm (enc)}} W_V,
\end{equation}
where $W_V\in \mathbb{R}^{d \times d_h} $ is learned parameter. Finally, similarly to \cite{pundak2018deep, chang2021context}, we concatenate \( D_{n} \) and $B_{n}$ to jointly determine the output of the next token $\hat{t}_n$. Therefore, \eqref{eq:1} becomes:
\begin{equation}
\label{eq:8}
\hat{t}_n = \text{Softmax}( (D_{n} \oplus B_{n}) W_n  + b_n).
\end{equation}




\subsection{Contrastive Entity Disambiguation (CED)}
\label{sec:ced}
Although context attention mechanisms help bias the decoder toward relevant entities, they often struggle to resolve phonetic ambiguities when multiple candidate entities have near-identical pronunciations. This is particularly problematic under noisy or uncertain acoustic conditions, where attention weights become diffused across several similar options. To address this challenge, we introduce a CED loss to explicitly enhance the discriminative capacity of the decoder’s hidden representations. 

Specifically, at each decoder step $n$, we define a contrastive objective that encourages the decoder output $D_{n}$ to be closer to the representation of the correct bias entity \( N^{\mathrm{(enc)}}_{p} \), while being distant from a set of $I$ phonetically similar but semantically incorrect hard negatives \( \{ N^{\mathrm{(enc)}}_{n_i} \}_{i=1}^{I} \subset N^{\mathrm{(enc)}},\ n_i \ne p \). The training loss is based on the InfoNCE formulation:
\vspace{-2mm}
\begin{equation}
\label{eq:9}
\mathcal{L}_{\text{CED}} = -\log \frac{e^{\text{sim}(D_{n}, N^{\mathrm{(enc)}}_{p})/\tau}}{e^{\text{sim}(D_{n}, N^{\mathrm{(enc)}}_{p})/\tau} + \sum_{i=1}^{I} e^{\text{sim}(D_{n}, N^{\mathrm{(enc)}}_{n_i})/\tau}},
\end{equation}
where \( \text{sim}(\cdot, \cdot) \) denotes cosine similarity, and \( \tau \) is a temperature hyperparameter that controls the sharpness of the distribution. The hard negative sampling strategy is described in Section~\ref{ssec:bias_con}.

\vspace{-1mm}
\subsection{Joint Training}

Unlike previous approaches \cite{tang2024improving, sudo2024contextualized}, where each token within an entity is assigned the same bias index to guide token-level alignment with biased phrases, our proposed entity loss adopts a more targeted strategy.
As illustrated in Fig.~\ref{fig:model}, consider the sentence:  
``\begin{CJK}{UTF8}{gbsn}苏炳添李金哲李玲等都受了一些影响\end{CJK}''  
(English translation: \textit{Su1 Bing3tian1, Li3 Jin1zhe2, Li3 Ling2, and others were all affected to some extent.})  
Suppose the biased entities ``\begin{CJK}{UTF8}{gbsn}苏炳添\end{CJK}'' (Su1 Bing3tian1, entity index \(C_1\)), ``\begin{CJK}{UTF8}{gbsn}李金哲\end{CJK}'' (Li3 Jin1zhe2, \(C_2\)), and ``\begin{CJK}{UTF8}{gbsn}李玲\end{CJK}'' (Li3 Ling2, \(C_5\)) are present in the biasing list.
The method in \cite{tang2024improving, sudo2024contextualized} assigns the same index to all tokens of each entity, resulting in the label sequence:
\vspace{-2mm}
\[
[C_1, C_1, C_1, C_2, C_2, C_2, C_5, C_5, C_0, C_0, \dots].
\]

In contrast, our method assigns the entity index only to the first token of each entity, while the remaining tokens are labeled as \(C_0\), yielding:
\[
[C_1, C_0, C_0, C_2, C_0, C_0, C_5, C_0, C_0, C_0, \dots].
\]

This design explicitly guides the model to retrieve the entire entity from the biasing list at the correct decoding step—i.e., the step where the first token is generated. From subsequent steps (e.g., decoding steps 2 and 3), no further selection is made. This mitigates the issue of incomplete decoding of multi-token entities caused by token-wise matching.
The entity loss is defined as:
\vspace{-1mm}
\begin{equation}
\mathcal{L}_{\text{Entity}} = -\sum_{n=1}^{N} \log p_c(\beta_n \mid T_{1:n-1}),
\end{equation}
where \(\beta_n \in (\beta_1, \beta_2, \dots, \beta_L)\) denotes the constructed entity index sequence.

The overall training objective combines four losses:
\begin{equation}
\label{eq:loss}
\mathcal{L} = \lambda \mathcal{L}_{\text{ASR}} + (1 - \lambda) \mathcal{L}_{\text{CTC}} + \mathcal{L}_{\text{Entity}} + \mathcal{L}_{\text{CED}},
\end{equation}
where \(\mathcal{L}_{\text{ASR}}\) and \(\mathcal{L}_{\text{CTC}}\) are standard ASR objectives, and \(\mathcal{L}_{\text{CED}}\) is the contrastive entity disambiguation loss described in Section~\ref{sec:ced}.

\vspace{-2mm}
\subsection{Hierarchical Entity Filtering (HEF)}
\label{sec:hef}
To improve the precision of entity selection during inference, we design a HEF strategy based on both phonetic similarity and a confidence threshold. This process refines the biasing list prior to computing the final prediction probabilities, reducing the chance of incorrect or irrelevant selecting.

\noindent \textbf{Phonetic-Aware Pre-selection.} Specifically, prior to decoding each step $n$, we compute the attention distribution 
$s_n \in \mathbb{R}^{L+1}$
over the biasing list as defined in \eqref{eq:6}, and identify the most attended entity by selecting the index with the maximum attention score:
\vspace{-1mm}
\begin{equation}
\hat{l} = \arg\max_{l} s_{n,l}.
\end{equation}

If $C_{\hat{l}} \neq C_0$ (i.e., not the “$<$no-bias$>$” token), we regard $C_{\hat{l}}$ as the dominant attended entity at step $n$. On the basis of this anchor, we retrieve its top-{K} phonemically similar entities from the biasing list using an edit-distance-based retrieval algorithm at the phoneme level. This enables the inclusion of acoustically confusable candidates (e.g., “陈冠希 (Chen2 Guan4Xi1, Person)” vs. “陈观鑫 (Chen2 Guan1Xin1, Person)”) as high-priority alternatives for contextual refinement.

We then construct a refined biasing list by selecting the representations of the retrieved phonemically similar entities along with the special “$<$no-bias$>$” token. Formally, we define the filtered biasing set as:

\begin{equation}
B_{n}^{\text{filtered}} = \{C_0\} \cup \{C_{l} \mid l \in \text{Top-}K(C_{\hat{l}})\},
\end{equation}
where $\text{Top-}K(C_{\hat{l}})$ denotes the indices of the top-{K} entities most similar to $C_{\hat{l}}$ in phoneme space. This filtered set $B_{n}^{\text{filtered}}$ is then used in place of the full biasing list for the subsequent attention computation in step $n$, thereby enabling more focused and acoustically-informed context modeling. Therefore, \eqref{eq:8} becomes:
\begin{equation}
\hat{t}_n = \text{Softmax}( (D_{n} \oplus B_{n}^{\text{filtered}}) W_n  + b_n).
\end{equation}


\noindent \textbf{Confidence-Based Gating.} Finally, to prevent unreliable entity biasing toward low-confidence entities, we incorporate a confidence-based gating mechanism during decoding. If the model’s highest selection probability over all entities in the filtered biasing set (excluding the “$<$no-bias$>$” token $C_0$) falls below a predefined threshold $\sigma$, the model defaults to generating from the standard token vocabulary. Formally, let:

\begin{equation}
\delta = \mathbb{I}\left[\max_{\beta_{n} \in B_{n}^{\text{filtered}}, \beta_{n} \ne C_0} p_c(\beta_{n} | T_{1:n-1}) < \sigma\right],
\end{equation}
where $\mathbb{I}[\cdot]$ denotes the indicator function that evaluates to 1 when the enclosed condition holds, and 0 otherwise. The final entity selection probability is then defined as:
\begin{equation}
p_c(\beta_{n} \mid T_{1:n-1}) =
\begin{cases}
1, & \text{if } \delta = 1 \text{ and } \beta_{n} = C_0 \\
0, & \text{if } \delta = 1 \text{ and } \beta_{n} \ne C_0 \\
p_c(\beta_{n} \mid T_{1:n-1}), & \text{if } \delta = 0
\end{cases}
\end{equation}


The proposed HEF strategy, which integrates phonetic-aware pre-selection and confidence-based gating, mitigates incorrect entity inclusion while retaining high recall.


\section{Experimental Evaluations}

\subsection{Implementation Details}
\label{ssec:Implementation Details}

All models were trained on 4 NVIDIA A100 80GB GPUs with a batch size of 64. The input consisted of 80-dimensional log-Mel filterbank features, extracted using a 25~ms window and 10~ms frame shift. These acoustic features were downsampled by a 2D convolutional front-end to one fourth of their original temporal resolution, followed by a linear projection to 256 dimensions.
The ASR backbone was a joint CTC/attention-based Conformer model \cite{gulati20_interspeech}, comprising 12 encoder layers and 4 attention heads, with an encoder output dimension of 256. The decoder contained 4 layers with 512-dimensional hidden states.
For text and phoneme encoding, we employed LSTM-based encoders consisting of 3 layers, each with 512 hidden units. The weight of the CTC loss, $\lambda$, was set to 0.7. 
The temperature hyperparameter \( \tau \) in \eqref{eq:9} was set to 0.1.
Based on comparative experiments, the size of the similar-entity pool $K$ and the threshold parameter $\sigma$ for the HEF strategy proposed in Section~\ref{sec:hef} were ultimately set to 20 and 0.9, respectively. 

\begin{table}[t]
\centering
\caption{Dataset statistics.  ``Utt." and ``NE" mean the number of utterances and named entities, respectively. For LibriSpeech, we use the test-clean subset.}
\vspace{-3mm}
\label{tab:dataset_statistics}
\begin{adjustbox}{max width=0.47\textwidth}
\begin{tabular}{l|cccccc}
\toprule
\multirow{2}{*}{\textbf{Dataset}} & \multicolumn{2}{c}{\textbf{Train}} & \multicolumn{2}{c}{\textbf{Dev}} & \multicolumn{2}{c}{\textbf{Test}} \\
\cmidrule(lr){2-3} \cmidrule(lr){4-5} \cmidrule(lr){6-7}
  & \textbf{Utt.} & \textbf{NE} & \textbf{Utt.} & \textbf{NE} & \textbf{Utt.} & \textbf{NE} \\
\midrule
\midrule
AISHELL-1     & 119919 & 14241 & 14326 & 2194 & 7176  & 1186 \\
DATA2     & 64570  & 11858 & 3100 & 2568 & 3100 & 2508 \\
THCHS-30         & --  & -- & --  & -- & 2495  & 268 \\
LibriSpeech          & --  & -- & --  & -- & 2620  & 2130 \\

\bottomrule
\end{tabular}
\end{adjustbox}
\vspace{-5mm}
\end{table}

\vspace{-1mm}
\subsection{Experimental Conditions}
\vspace{-1mm}
\label{ssec:Experiment Conditions}
\noindent \textbf{Datasets.} We evaluated the effectiveness of the proposed method PARCO on the Chinese AISHELL-1 \cite{bu2017aishell} and English DATA2 dataset \cite{yadav2020end}.
Additionally, to verify PARCO's generalization ability, we also introduced two test sets of the Chinese THCHS-30 \cite{wang2015thchs} and English LibriSpeech test-clean dataset \cite{2015LibriSpeech}. Table \ref{tab:dataset_statistics} provides detailed statistics.

\noindent \textbf{Evaluation Metrics.} 
For Chinese datasets, we evaluate ASR performance using character error rate (CER) and assess NE recognition accuracy with NE-CER, following prior work \cite{han2021cif, zhou2023copyne}. For English datasets, we use word error rate (WER) and NE-WER to measure overall and entity-level performance, respectively.

\noindent \textbf{Baselines.} 
In the experiments, we compared three baselines with our proposed PARCO:
\begin{itemize}[leftmargin=*]
\setlength{\itemsep}{0pt}
\setlength{\parsep}{0pt}
\setlength{\parskip}{0pt}

\item \textbf{CBA} \cite{zhang2022end} introduces a contextual bias attention module to E2E ASR models, which attends to a bias phrase embedding using decoder-side attention context. This allows the model to adapt its output distribution to better recognize infrequent contextual entities.

\item \textbf{CopyNE} \cite{zhou2023copyne} addresses the challenge of NE recognition in ASR by copying entities from a preloaded NE dictionary, enabling complete and accurate entity transcription.

\item \textbf{ED-CEC} \cite{he2023ed} introduces a contextual ASR postprocessing method that enhances the recognition of rare words by detecting errors and using context-aware correction, optimizing the decoding process and leveraging a rare word list for improved accuracy.


\end{itemize}

\begin{table*}[t]
\caption{Performance on AISHELL-1 and DATA2 under different numbers of distractors $N$. Each cell shows CER/NE-CER or WER/NE-WER. Values in parentheses denote relative NE-CER or NE-WER reduction (\%).}
\vspace{-3mm}
\label{tab:results}
\centering
\begin{adjustbox}{max width=\textwidth}
\begin{tabular}{l|c|c|c|c}
\toprule
\textbf{Model} & \textbf{$N$=0} & \textbf{$N$=100} & \textbf{$N$=1000} & \textbf{$N$=5000} \\
\midrule
\midrule
\rowcolor[HTML]{EFEFEF}
\multicolumn{5}{c}{\textit{AISHELL-1 (CER $\downarrow$ (CERR $\uparrow$) / NE-CER $\downarrow$ (NE-CERR $\uparrow$))}} \\
\midrule
Conformer \cite{gulati20_interspeech} & 4.94 (--) / 9.60 (--) & 4.94 (--) / 9.60 (--) & 4.94 (--) / 9.60 (--) & 4.94 (--) / 9.60 (--) \\
\hspace{1em} + CBA \cite{zhang2022end}    & 4.57 (+7.49) / 5.36 (+44.17) & 4.65 (+5.87) / 5.91 (+38.44) & 4.86 (+1.62) / 6.82 (+28.96) & 5.18 (-4.86) / 8.12 (+15.42) \\
\hspace{1em} + CopyNE \cite{zhou2023copyne} & 4.37 (+11.54) / 2.24 (+76.67) & 4.48 (+9.31) / 2.97 (+69.06) & 4.72 (+4.45) / 3.76 (+60.83) & 5.05 (-2.23) / 4.80 (+50.00) \\
\hspace{1em} + ED-CEC \cite{he2023ed}  & 4.49 (+9.11) / 4.96 (+48.33) & 4.54 (+8.10) / 5.23 (+45.52) & 4.66 (+5.67) / 5.68 (+40.83) & 4.81 (+2.63) / 6.31 (+34.27) \\
\midrule
\rowcolor[HTML]{D9F0D3}
\hspace{1em} + \textbf{PARCO} & \textbf{4.03 (+18.42) / 1.57 (+83.65)} & \textbf{4.04 (+18.22) / 1.69 (+82.40)} & \textbf{4.22 (+14.57) / 2.84 (+70.42)} & \textbf{4.45 (+9.92) / 3.56 (+62.92)} \\
\midrule
\midrule
\rowcolor[HTML]{EFEFEF}
\multicolumn{5}{c}{\textit{DATA2 (WER $\downarrow$ (WERR $\uparrow$) / NE-WER $\downarrow$ (NE-WERR $\uparrow$))}} \\
\midrule
Conformer \cite{gulati20_interspeech} & 12.05 (--) / 26.64 (--) & 12.05 (--) / 26.64 (--) & 12.05 (--) / 26.64 (--) & 12.05 (--) / 26.64 (--) \\
\hspace{1em} + CBA \cite{zhang2022end}    & 11.12 (+7.72) / 18.78 (+29.50) & 11.52 (+4.40) / 20.71 (+22.26) & 12.28 (-1.91) / 25.46 (+4.43) & 13.23 (-9.79) / 28.04 (-5.26) \\
\hspace{1em} + CopyNE \cite{zhou2023copyne}  & 10.26 (+14.85) / 10.34 (+61.19) & 10.54 (+12.53) / 12.57 (+52.82) & 11.44 (+5.06) / 19.14 (+28.15) & 12.31 (-2.16) / 26.60 (+0.15) \\
\hspace{1em} + ED-CEC \cite{he2023ed}  & 10.59 (+12.12) / 16.54 (+37.91) & 10.66 (+11.54) / 16.87 (+36.67) & 11.38 (+5.56) / 17.94 (+32.66) & 11.67 (+3.15) / 20.76 (+22.07) \\
\midrule
\rowcolor[HTML]{D9F0D3}
\hspace{1em} + \textbf{PARCO} & \textbf{10.00 (+17.01) / 8.34 (+68.69)} & \textbf{10.17 (+15.60) / 9.47 (+64.45)} & \textbf{11.14 (+7.55) / 14.16 (+46.85)} & \textbf{11.49 (+4.65) / 17.15 (+35.62)} \\
\bottomrule
\end{tabular}
\end{adjustbox}
\vspace{-5mm}
\end{table*}

\subsection{Biasing List Construction}
\vspace{-1mm}
\label{ssec:bias_con}
\noindent \textbf{Entity Selection.} 
For AISHELL-1, we adopted the named entities provided by Chen et al. \cite{chen2022aishell}. For THCHS-30, we extracted person, location, and organization using \texttt{HanLP}\footnote{\href{https://github.com/hankcs/HanLP}{https://github.com/hankcs/HanLP}}. For DATA2, we utilized the annotated entities—person, location, and organization—available in \cite{yadav2020end}. For LibriSpeech, we employed \texttt{NuNER\_Zero}\footnote{\href{https://huggingface.co/numind/NuNER\_Zero}{https://huggingface.co/numind/NuNER\_Zero}} to extract the same three types of entities.
During training, following prior work \cite{huber2021instant, zhou2023copyne}, we randomly selected two- or three-character substrings (or two- or three-word spans in English) from non-entity utterances and treated them as entities if they contained relevant phoneme sequences. Phoneme sequences for the Chinese and English datasets were generated using \texttt{pypinyin}\footnote{\href{https://pypi.org/project/pypinyin}{https://pypi.org/project/pypinyin}} and \texttt{g2pE}\footnote{\href{https://github.com/Kyubyong/g2p}{https://github.com/Kyubyong/g2p}}, respectively.

\noindent\textbf{Hard Negative Sampling.}
To construct meaningful negative examples during training, we select 1–3 hard negative entities for each ground-truth (GT) entity by retrieving entries from the training set that exhibit the most similar pronunciations. Specifically, we compute phoneme-level edit distances between the target entity and all other entities in the training vocabulary, then choose the top nearest ones that differ in meaning. This ensures that the distractors are phonetically confusable yet semantically incorrect. For example, as shown in Fig. \ref{fig:model}, the target entity “李玲 (Li3 Ling2, Person)” may be paired with “李宁 (Li3 Ning2, Person)”, “李琳 (Li3 Lin2, Person)”, and “李滢 (Li3 Ying2, Person)” as hard negatives. 
To improve the model’s ability to distinguish true entities from distractors in large biasing lists, we include these hard negatives in the training objective alongside GT entities. During inference, we construct the biasing list by combining GT entities with a set of phonemically similar distractors for each utterance.

\vspace{-1mm}
\subsection{Results and Analysis}
\vspace{-1mm}
\label{ssec:Results and Analysis}

\noindent \textbf{Comparative Results under Varying Biasing List Sizes.}
As shown in Table~\ref{tab:results}, PARCO consistently outperforms all baselines across different distractor settings on both AISHELL-1 and DATA2. On AISHELL-1, it achieves the lowest NE-CER in all cases—1.57\%, 1.69\%, 2.84\%, and 3.56\% at $N=0, 100, 1,000, 5,000$, respectively. Similarly, on DATA2, it yields the lowest NE-WER under each setting—8.34\%, 9.47\%, 14.16\%, and 17.15\%.
Although increasing the number of distractors generally leads to performance degradation for all models, PARCO exhibits stronger robustness. For example, compared with the strongest baseline ED-CEC, PARCO maintains lower NE-CERs and NE-WERs even at large $N$, highlighting its better ability to resist noise from confusable entities. We attribute this improvement to the proposed HEF strategy, which helps narrow down candidate sets via phonetic similarity and confidence thresholding.

\noindent \textbf{Ablation Study.} To assess the contribution of each module in PARCO, we conduct an ablation study under the most challenging setting with 5,000 distractors. As shown in Table~\ref{tab:ablation}, removing the HEF strategy leads to noticeable performance drops on both AISHELL-1 and DATA2, confirming its effectiveness in reducing false entity predictions.
Further, omitting the CED loss or the entity loss also causes consistent degradation, highlighting the necessity of both objective terms in supervising entity-specific learning.
We observe more severe performance drops when removing either the phoneme encoder or the text encoder, indicating that both phonetic and contextual representations are indispensable to accurate entity prediction.
Overall, the results validate the design of PARCO and demonstrate that each component contributes to its final performance.

\begin{table}[t]
\caption{Ablation study on AISHELL-1 and DATA2 (\%). NE-CER / NE-WER are reported with relative error rate reduction ($\uparrow$) in parentheses under 5,000 distractors. ``w/o": ``without", TE: Text Encoder, PE: Phoneme Encoder, CED: Contrastive Entity Disambiguation, HEF: Hierarchical Entity Filtering.}
\vspace{-3mm}
\centering
\label{tab:ablation}
\begin{adjustbox}{max width=0.47\textwidth}
\begin{tabular}{l|l|c|c}
\toprule
\textbf{No.} & \textbf{Configuration} & \textbf{AISHELL-1} & \textbf{DATA2} \\
\midrule
\midrule
\rowcolor[HTML]{D9F0D3}
\textbf{1} & \textbf{PARCO}  & \textbf{3.56 (+62.92)} & \textbf{17.15 (+35.62)} \\
2 & No.1 w/o HEF & 4.07 (+57.60) & 19.52 (+26.73) \\
3 & No.2 w/o CED Loss & 4.39 (+54.27) & 20.10 (+24.55) \\
4 & No.3 w/o Entity Loss & 5.51 (+42.60) & 21.25 (+20.23) \\
5 & No.4 w/o PE & 8.27 (+13.85) & 28.15 (-5.67) \\
6 & No.5 w/o TE & 9.60 (--) & 26.64 (--) \\
\bottomrule
\end{tabular}
\end{adjustbox}
\vspace{-5mm}
\end{table}

\begin{table}[t]
\caption{
Robustness analysis on OOD datasets (\%). 
Values in parentheses denote the relative reduction compared with the Conformer baseline.
}
\vspace{-3mm}
\centering
\label{tab:robustness}
\begin{adjustbox}{max width=0.47\textwidth}
\begin{tabular}{l|cccc}
\toprule
\multirow{2}{*}{\textbf{Model}} & \multicolumn{2}{c}{\textbf{THCHS-30}} & \multicolumn{2}{c}{\textbf{LibriSpeech test-clean}}  \\
\cmidrule{2-5}
 & CER $\downarrow$  & NE-CER $\downarrow$ & WER $\downarrow$ & NE-WER $\downarrow$ \\
\midrule
\midrule
Conformer \cite{gulati20_interspeech} 
& 27.21 (--) & 46.33 (--) & 7.91 (--) & 12.39 (--) \\
\hspace{1em} + CBA \cite{zhang2022end} 
& 28.55 (-4.92) & 47.61 (-2.76) & 9.68 (-22.38) & 13.72 (-10.73) \\
\hspace{1em} + CopyNE \cite{zhou2023copyne} 
& 25.29 (+7.06) & 15.65 (+66.22) & 9.31 (-17.70) & 10.93 (+11.78) \\
\hspace{1em} + ED-CEC \cite{he2023ed} 
& 30.11 (-10.66) & 50.52 (-9.04) & 8.62 (-8.98) & 14.38 (-16.06) \\
\midrule
\rowcolor[HTML]{D9F0D3} 
\hspace{1em} \textbf{+ PARCO} 
& \textbf{21.53 (+20.87)} 
& \textbf{8.48 (+81.70)}  
& \textbf{7.09 (+10.37)} 
& \textbf{6.17 (+50.20)} \\
\bottomrule
\end{tabular}
\end{adjustbox}
\vspace{-3mm}
\end{table}

\noindent \textbf{Robustness Analysis.} 
We evaluate the robustness of models via cross-domain testing. Specifically, we apply the model trained on the Chinese dataset AISHELL-1 to the OOD dataset THCHS-30, and the model trained on the English dataset DATA2 to LibriSpeech. 
As shown in Table~\ref{tab:robustness}, baseline models suffer from high NE-CER/NE-WER on unseen data, indicating their vulnerability to domain shift. In contrast, our proposed PARCO achieves the lowest error rates across both datasets, reducing NE-CER by 81.70\% on THCHS-30 and NE-WER by 50.20\% on LibriSpeech compared with the base Conformer. These results demonstrate the strong generalization ability of our entity-aware correction framework.



\begin{figure}[t]
\centerline{\includegraphics[width=1\columnwidth]{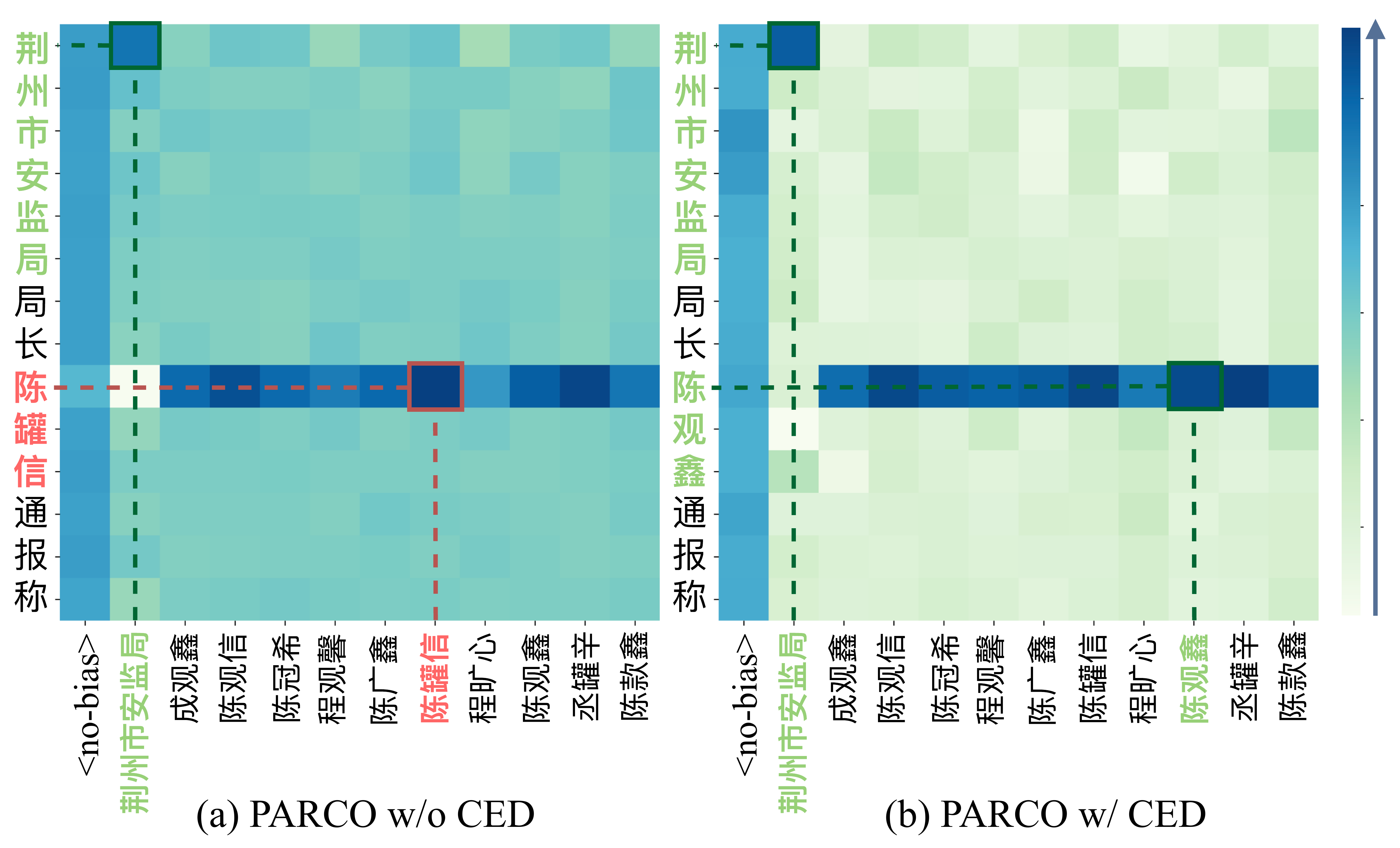}}
\vspace{-3mm}
\caption{Comparison of attention visualization with and without CED loss. The horizontal and vertical axes are the biasing list and the transcript, respectively.}
\label{fig:CED_loss}
\vspace{-6mm}
\end{figure}

\noindent \textbf{Qualitative Analysis.}
To verify the effectiveness of the proposed CED loss, we construct a challenging test case on AISHELL-1 consisting of 10 candidate entities, including the correct bias entity \textcolor{teal}{陈观鑫 (Chen2 Guan1Xin1, Person)} and 9 hard negative distractors that are phonetically similar but semantically incorrect. These distractors are created by altering initials, finals, or nasal endings to introduce fine-grained phonetic confusion. The 9 distractors include: 
\textcolor{Maroon}{成观鑫 (Cheng2 Guan1Xin1, Person)}, 
\textcolor{Maroon}{陈观信 (Chen2 Guan1Xin4, Person)},
\textcolor{Maroon}{陈冠希 (Chen2 Guan4Xi1, Person)},
\textcolor{Maroon}{程观馨 (Cheng2 Guan1Xin1, Person)},
\textcolor{Maroon}{陈广鑫 (Chen2 Guang3Xin1, Person)},
\textcolor{Maroon}{陈罐信 (Chen2 Guan4Xin4, Person)},
\textcolor{Maroon}{程旷心 (Cheng2 Kuang4Xin1, Person)},
\textcolor{Maroon}{丞罐辛 (Cheng2 Guan4Xin1, Person)},
\textcolor{Maroon}{陈款鑫 (Chen2 Kuan3Xin1, Person)}.
Fig.~\ref{fig:CED_loss} compares the attention distribution of the decoder over the biasing list before and after applying the CED loss. In the subfigure (a), without CED, the model incorrectly allocates high attention weights to the hard negative entity “\textcolor{Maroon}{陈罐信}”, which is phonetically similar to the correct target “\textcolor{teal}{陈观鑫}”. This misalignment arises from the model’s limited ability to disambiguate phonetically similar entities, especially under noisy or uncertain acoustic conditions. In contrast, the subfigure (b) shows that with CED training, the attention distribution becomes more focused and discriminative. The decoder correctly attends to the intended entity “\textcolor{teal}{陈观鑫}”, suppressing attention to similar distractors. This demonstrates that our contrastive training objective effectively guides the decoder to learn fine-grained semantic distinctions between phonetically confusable entities, improving its robustness in bias-aware recognition tasks.

Table~\ref{tab:examples} further shows representative transcription examples on AISHELL-1 and DATA2. Although baseline models either hallucinate entities or only partially recover names, PARCO generates complete and correct entity spans in both Chinese and English contexts. This highlights PARCO’s advantage in enforcing entity integrity and resisting confusion from phonetically similar or out-of-context entities.

\begin{table}[t]
\centering
\renewcommand{\arraystretch}{1.5}
\setlength{\tabcolsep}{4pt}
\caption{Comparative examples of transcriptions generated by different models. \textcolor{Maroon}{Red text} indicates errors, while \textcolor{teal}{green text} highlights correctly transcribed entities.}
\vspace{-2mm}

\begin{tabular}{|>{\raggedright\arraybackslash}m{1.8cm}|>{\raggedright\arraybackslash}m{6.0cm}|}
\hline
\multicolumn{2}{|c|}{\cellcolor[HTML]{EFEFEF}\textbf{AISHELL-1}} \\
\hline
\cellcolor[HTML]{D9EAD3}\textbf{Entities:} & \cellcolor[HTML]{D9EAD3}\centering 罗宾索尔科维 (\textbf{Robin Szolkowy, Person}), \\ 罗宾斯切姆贝拉 (\textbf{Robin Schembera, Person}) \tabularnewline
\hline
\textbf{Translation} & \textit{800 meters: Robin Schembera} \\
\textbf{Ground Truth} & 八百米罗宾斯切姆贝拉 \\
\textbf{CBA} \cite{zhang2022end} & 八百米\textcolor{teal}{罗宾斯}\textcolor{Maroon}{窃}\textcolor{teal}{姆贝拉} \\
\textbf{CopyNE} \cite{zhou2023copyne} & 八百米\textcolor{teal}{罗}\textcolor{Maroon}{滨}\textcolor{teal}{斯切}\textcolor{Maroon}{尔}\textcolor{teal}{姆贝拉} \\
\textbf{ED-CEC} \cite{he2023ed} & 八百米\textcolor{teal}{罗宾}\textcolor{Maroon}{索尔科维姆被}\textcolor{teal}{拉} \\
\hline
\textbf{PARCO} & 八百米\textcolor{teal}{罗宾斯切姆贝拉}\\
\hline
\hline
\multicolumn{2}{|c|}{\cellcolor[HTML]{EFEFEF}\textbf{DATA2}} \\
\hline
\cellcolor[HTML]{D9EAD3}\textbf{Entities:} & \cellcolor[HTML]{D9EAD3}\centering WM WALTERS CARPENTER'S, WM JONES \tabularnewline
\hline
\textbf{Ground Truth} & WM WALTERS CARPENTER'S CREW WM JONES DOG DRIVER \\
\textbf{CBA} \cite{zhang2022end} & \textcolor{Maroon}{WILLIAM WALTIS CARPING HIS} CREW \textcolor{Maroon}{WILLIAM} \textcolor{teal}{JONES} DON'T DRIVER \\
\textbf{CopyNE} \cite{zhou2023copyne} & \textcolor{Maroon}{WILLIAM WALTIS CARPENTIS} CREW \textcolor{Maroon}{WILLIAM} \textcolor{teal}{JONES} DOG DRIVER \\
\textbf{ED-CEC} \cite{he2023ed} & \textcolor{Maroon}{WILLIAM} \textcolor{teal}{WALTERS CARPENTER'S} CREW \textcolor{Maroon}{WILLIAM} \textcolor{teal}{JONES} DOG DRIVER \\
\hline
\textbf{PARCO} & \textcolor{teal}{WM WALTERS CARPENTER'S} CREW \textcolor{teal}{WM JONES} DOG DRIVER \\
\hline
\end{tabular}
\label{tab:examples}
\vspace{-5mm}
\end{table}

\vspace{-1mm}
\section{Conclusion}
In this paper, we introduce PARCO, a phoneme-augmented contextual ASR framework that addresses the challenges of homophone disambiguation and incomplete multi-token entity recognition. PARCO integrates four key components: phoneme-aware encoding to capture fine-grained pronunciation differences, contrastive entity disambiguation to enhance discriminative decoding, entity-level supervision to ensure span-level integrity, and hierarchical entity filtering to improve retrieval precision and robustness. Experimental results show that PARCO significantly improves contextual biasing performance on both Chinese (AISHELL-1) and English (DATA2) benchmarks, outperforming strong baselines. Moreover, it generalizes well to out-of-domain datasets such as THCHS-30 and LibriSpeech. In future work, we plan to extend PARCO to multilingual settings and explore its integration with retrieval-augmented speech understanding systems.
\end{CJK}

\section*{Acknowledgment}
This work was partly supported by JST CREST Grant Number JPMJCR22D1, Japan.

\clearpage
\bibliographystyle{IEEEtran}
\bibliography{IEEEtran}

\begin{thebibliography}{10}
\providecommand{\url}[1]{#1}
\csname url@samestyle\endcsname
\providecommand{\newblock}{\relax}
\providecommand{\bibinfo}[2]{#2}
\providecommand{\BIBentrySTDinterwordspacing}{\spaceskip=0pt\relax}
\providecommand{\BIBentryALTinterwordstretchfactor}{4}
\providecommand{\BIBentryALTinterwordspacing}{\spaceskip=\fontdimen2\font plus
\BIBentryALTinterwordstretchfactor\fontdimen3\font minus \fontdimen4\font\relax}
\providecommand{\BIBforeignlanguage}[2]{{%
\expandafter\ifx\csname l@#1\endcsname\relax
\typeout{** WARNING: IEEEtran.bst: No hyphenation pattern has been}%
\typeout{** loaded for the language `#1'. Using the pattern for}%
\typeout{** the default language instead.}%
\else
\language=\csname l@#1\endcsname
\fi
#2}}
\providecommand{\BIBdecl}{\relax}
\BIBdecl

\bibitem{kheddar2024automatic}
H.~Kheddar, M.~Hemis, and Y.~Himeur, ``Automatic speech recognition using advanced deep learning approaches: A survey,'' \emph{Information Fusion}, pp. 102\,422--102\,441, 2024.

\bibitem{zhang2024cif}
T.-H. Zhang, D.~Zhou, G.~Zhong, J.~Zhou, and B.~Li, ``{CIF-T}: A novel {CIF}-based transducer architecture for automatic speech recognition,'' in \emph{Proc. ICASSP}, 2024, pp. 10\,531--10\,535.

\bibitem{xu2024dynamic}
J.~Xu, W.~Zhou, Z.~Yang, E.~Beck, and R.~Schl{\"u}ter, ``Dynamic encoder size based on data-driven layer-wise pruning for speech recognition,'' in \emph{Proc. Interspeech}, 2024, pp. 4563--4567.

\bibitem{huber2021instant}
C.~Huber, J.~Hussain, S.~St{\"u}ker, and A.~Waibel, ``Instant one-shot word-learning for context-specific neural sequence-to-sequence speech recognition,'' in \emph{Proc. {ASR}U}, 2021, pp. 1--7.

\bibitem{yang2024multi}
Z.~Yang, J.~He, and T.~Toda, ``Multi-modal video summarization based on two-stage fusion of audio, visual, and recognized text information,'' in \emph{Proc. APSIPA ASC}, 2024, pp. 1--6.

\bibitem{he24_interspeech}
J.~He and T.~Toda, ``{2DP-2MRC}: 2-dimensional pointer-based machine reading comprehension method for multimodal moment retrieval,'' in \emph{Proc. Interspeech}, 2024, pp. 5073--5077.

\bibitem{he2024mf}
J.~He, X.~Shi, X.~Li, and T.~Toda, ``{MF-AED-AEC}: Speech emotion recognition by leveraging multimodal fusion, {{ASR}} error detection, and {{ASR}} error correction,'' in \emph{Proc. ICASSP}, 2024, pp. 11\,066--11\,070.

\bibitem{li2024speech}
Y.~Li, P.~Bell, and C.~Lai, ``Speech emotion recognition with {ASR} transcripts: A comprehensive study on word error rate and fusion techniques,'' \emph{arXiv:2406.08353}, 2024.

\bibitem{shi2024study}
X.~Shi, Y.~Gao, J.~He, J.~Mi, X.~Li, and T.~Toda, ``A study on multimodal fusion and layer adapter in emotion recognition,'' in \emph{Proc. APSIPA ASC}, 2024, pp. 1--6.

\bibitem{tian2023semi}
J.~Tian, D.~Hu, X.~Shi, J.~He, X.~Li, Y.~Gao, T.~Toda, X.~Xu, and X.~Hu, ``Semi-supervised multimodal emotion recognition with consensus decision-making and label correction,'' in \emph{Proc. MRAC}, 2023, pp. 67--73.

\bibitem{mi2024two}
J.~Mi, X.~Shi, D.~Ma, J.~He, T.~Fujimura, and T.~Toda, ``Two-stage framework for robust speech emotion recognition using target speaker extraction in human speech noise conditions,'' in \emph{Proc. APSIPA ASC}, 2024.

\bibitem{he25_interspeech}
J.~He, J.~Mi, and T.~Toda, ``{GIA-MIC}: Multimodal emotion recognition with gated interactive attention and modality-invariant learning constraints,'' in \emph{Proc. Interspeech}, 2025, pp. 1--5.

\bibitem{sun2022tree}
G.~Sun \emph{et~al.}, ``Tree-constrained pointer generator with graph neural network encodings for contextual speech recognition,'' in \emph{Proc. Interspeech}, 2022, pp. 2043--2047.

\bibitem{yang2024mala}
G.~Yang, Z.~Ma, F.~Yu, Z.~Gao, S.~Zhang, and X.~Chen, ``{MaLa-ASR}: Multimedia-assisted {LLM}-based {ASR},'' in \emph{Proc. Interspeech}, 2024, pp. 2405--2409.

\bibitem{huang2024improving}
R.~Huang, M.~Yarmohammadi, S.~Khudanpur, and D.~Povey, ``Improving neural biasing for contextual speech recognition by early context injection and text perturbation,'' in \emph{Proc. Interspeech}, 2024, pp. 752--756.

\bibitem{williams2018contextual}
I.~Williams, A.~Kannan, P.~S. Aleksic, D.~Rybach, and T.~N. Sainath, ``Contextual speech recognition in end-to-end neural network systems using beam search.'' in \emph{Proc. Interspeech}, 2018, pp. 2227--2231.

\bibitem{le2021deep}
D.~Le, G.~Keren, J.~Chan, J.~Mahadeokar, C.~Fuegen, and M.~L. Seltzer, ``Deep shallow fusion for {RNN-T} personalization,'' in \emph{Proc. SLT}, 2021, pp. 251--257.

\bibitem{chen2022factorized}
X.~Chen, Z.~Meng, S.~Parthasarathy, and J.~Li, ``Factorized neural transducer for efficient language model adaptation,'' in \emph{Proc. ICASSP}, 2022, pp. 8132--8136.

\bibitem{he2023ed}
J.~He, Z.~Yang, and T.~Toda, ``{ED-CEC}: Improving rare word recognition using {ASR} postprocessing based on error detection and context-aware error correction,'' in \emph{Proc. {ASR}U}, 2023, pp. 1--6.

\bibitem{li2024crossmodal}
Y.~Li, P.~Chen, P.~Bell, and C.~Lai, ``Crossmodal {ASR} error correction with discrete speech units,'' \emph{arXiv:2405.16677}, 2024.

\bibitem{he2023enhancing}
J.~He, Z.~Yang, and T.~Toda, ``Enhancing recognition of rare words in {ASR} through error detection and context-aware error correction,'' \emph{IEICE Tech. Rep.}, vol. 123, no. 292, pp. 13--18, 2023.

\bibitem{he2025pmf}
J.~He and T.~Toda, ``{PMF-CEC}: Phoneme-augmented multimodal fusion for context-aware asr error correction with error-specific selective decoding,'' \emph{IEEE Transactions on Audio, Speech and Language Processing}, vol.~33, pp. 2402--2417, 2025.

\bibitem{pundak2018deep}
G.~Pundak, T.~N. Sainath, R.~Prabhavalkar, A.~Kannan, and D.~Zhao, ``Deep context: end-to-end contextual speech recognition,'' in \emph{Proc. SLT}, 2018, pp. 418--425.

\bibitem{han2022improving}
M.~Han, L.~Dong, Z.~Liang, M.~Cai, S.~Zhou, Z.~Ma, and B.~Xu, ``Improving end-to-end contextual speech recognition with fine-grained contextual knowledge selection,'' in \emph{Proc. ICASSP}, 2022, pp. 8532--8536.

\bibitem{liu2022internal}
Y.~Liu, R.~Ma, H.~Xu, Y.~He, Z.~Ma, and W.~Zhang, ``Internal language model estimation through explicit context vector learning for attention-based encoder-decoder {ASR},'' in \emph{Proc. Interspeech}, 2022, pp. 1666--1670.

\bibitem{zhang2022end}
Z.~Zhang and P.~Zhou, ``End-to-end contextual {ASR} based on posterior distribution adaptation for hybrid {CTC}/attention system,'' \emph{arXiv:2202.09003}, 2022.

\bibitem{munkhdalai2023nam+}
T.~Munkhdalai, Z.~Wu, G.~Pundak, K.~C. Sim, J.~Li, P.~Rondon, and T.~N. Sainath, ``{NAM+}: Towards scalable end-to-end contextual biasing for adaptive {ASR},'' in \emph{Proc. SLT}, 2023, pp. 190--196.

\bibitem{zhou2023copyne}
S.~Zhou, Z.~Li, Y.~Hong, M.~Zhang, Z.~Wang, and B.~Huai, ``{CopyNE}: Better contextual {ASR} by copying named entities,'' in \emph{Proc. ACL}, 2024, pp. 2675--2686.

\bibitem{fang2025improving}
M.~Fang, T.~Wei, K.~Guo, Z.~Zhuang, Y.~Shi, N.~Cheng, S.~Wang, and J.~Xiao, ``Improving contextual asr with enhanced phrase-level representation based on mctc loss,'' in \emph{Proc. ICASSP}, 2025, pp. 1--5.

\bibitem{fang2025token}
M.~Fang, K.~Guo, T.~Wei, Z.~Zhuang, Y.~Shi, N.~Cheng, S.~Wang, and J.~Xiao, ``Token-level contextual network with ladder-shaped attention for end-to-end asr,'' in \emph{Proc. ICASSP}, 2025, pp. 1--5.

\bibitem{bruguier2019phoebe}
A.~Bruguier, R.~Prabhavalkar, G.~Pundak, and T.~N. Sainath, ``Phoebe: Pronunciation-aware contextualization for end-to-end speech recognition,'' in \emph{Proc. ICASSP}, 2019, pp. 6171--6175.

\bibitem{chen2019joint}
Z.~Chen, M.~Jain, Y.~Wang, M.~L. Seltzer, and C.~Fuegen, ``Joint grapheme and phoneme embeddings for contextual end-to-end {ASR}.'' in \emph{Proc. Interspeech}, 2019, pp. 3490--3494.

\bibitem{pandey2023procter}
R.~Pandey, R.~Ren, Q.~Luo, J.~Liu, A.~Rastrow, A.~Gandhe, D.~Filimonov, G.~Strimel, A.~Stolcke, and I.~Bulyko, ``{PROCTER}: Pronunciation-aware contextual adapter for personalized speech recognition in neural transducers,'' in \emph{Prco. ICASSP}, 2023, pp. 1--5.

\bibitem{futami2024phoneme}
H.~Futami, E.~Tsunoo, Y.~Kashiwagi, H.~Ogawa, S.~Arora, and S.~Watanabe, ``Phoneme-aware encoding for prefix-tree-based contextual {ASR},'' in \emph{Proc. ICASSP}, 2024, pp. 10\,641--10\,645.

\bibitem{kim2017joint}
S.~Kim, T.~Hori, and S.~Watanabe, ``Joint {CTC}-attention based end-to-end speech recognition using multi-task learning,'' in \emph{Proc. ICASSP}, 2017, pp. 4835--4839.

\bibitem{moriya2020self}
T.~Moriya, T.~Ochiai, S.~Karita, H.~Sato, T.~Tanaka, T.~Ashihara, R.~Masumura, Y.~Shinohara, and M.~Delcroix, ``Self-distillation for improving {CTC}-transformer-based {ASR} systems.'' in \emph{Proc. Interspeech}, 2020, pp. 546--550.

\bibitem{chang2021context}
F.-J. Chang, J.~Liu, M.~Radfar, A.~Mouchtaris, M.~Omologo, A.~Rastrow, and S.~Kunzmann, ``Context-aware transformer transducer for speech recognition,'' in \emph{Proc. {ASR}U}, 2021, pp. 503--510.

\bibitem{tang2024improving}
J.~Tang, K.~Kim, S.~Shon, F.~Wu, and P.~Sridhar, ``Improving {ASR} contextual biasing with guided attention,'' in \emph{Proc. ICASSP}, 2024, pp. 12\,096--12\,100.

\bibitem{sudo2024contextualized}
Y.~Sudo, M.~Shakeel, Y.~Fukumoto, Y.~Peng, and S.~Watanabe, ``Contextualized automatic speech recognition with attention-based bias phrase boosted beam search,'' in \emph{Proc. ICASSP}, 2024, pp. 10\,896--10\,900.

\bibitem{gulati20_interspeech}
A.~Gulati, J.~Qin, C.-C. Chiu, N.~Parmar, Y.~Zhang, J.~Yu, W.~Han, S.~Wang, Z.~Zhang, Y.~Wu, and R.~Pang, ``Conformer: Convolution-augmented transformer for speech recognition,'' in \emph{Proc. Interspeech}, 2020, pp. 5036--5040.

\bibitem{bu2017aishell}
H.~Bu, J.~Du, X.~Na, B.~Wu, and H.~Zheng, ``{AISHELL-1}: An open-source mandarin speech corpus and a speech recognition baseline,'' in \emph{Proc. O-COCOSDA}, 2017, pp. 1--5.

\bibitem{yadav2020end}
H.~Yadav, S.~Ghosh, Y.~Yu, and R.~R. Shah, ``End-to-end named entity recognition from english speech,'' in \emph{Proc. Interspeech}, 2020, pp. 4268--4272.

\bibitem{wang2015thchs}
D.~Wang and X.~Zhang, ``{THCHS}-30: A free {C}hinese speech corpus,'' \emph{arXiv:1512.01882}, 2015.

\bibitem{2015LibriSpeech}
V.~Panayotov, G.~Chen, D.~Povey, and S.~Khudanpur, ``{LibriSpeech}: An {ASR} corpus based on public domain audio books,'' in \emph{Proc. ICASSP}, 2015, pp. 5206--5210.

\bibitem{han2021cif}
M.~Han, L.~Dong, S.~Zhou, and B.~Xu, ``{CIF}-based collaborative decoding for end-to-end contextual speech recognition,'' in \emph{Proc. ICASSP}, 2021, pp. 6528--6532.

\bibitem{chen2022aishell}
B.~Chen, G.~Xu, X.~Wang, P.~Xie, M.~Zhang, and F.~Huang, ``{AISHELL-NER}: Named entity recognition from {C}hinese speech,'' in \emph{Proc. ICASSP}, 2022, pp. 8352--8356.

\end{thebibliography}

\end{document}